\documentclass[11pt, a4paper, copyright]{google}

\usepackage[authoryear, sort&compress, round]{natbib}
\makeatletter
\renewcommand{\absfont}{\normalfont\linespread{1.2}\fontsize{11}{12}\selectfont}
\makeatother
\definecolor{linkblue}{RGB}{0,0,139}      % 深蓝
\definecolor{navy}{RGB}{0,0,128}          % 海军蓝
\definecolor{royalblue}{RGB}{65,105,225}  % 皇家蓝
\definecolor{steelblue}{RGB}{70,130,180}  % 钢蓝
\definecolor{dodgerblue}{RGB}{30,144,255} % 道奇蓝
\definecolor{mediumblue}{RGB}{0,0,205}    % 中蓝
\definecolor{darkslateblue}{RGB}{72,61,139} % 深岩蓝
\usepackage[colorlinks = true,
            linkcolor = linkblue,
            urlcolor  = linkblue,
            citecolor = linkblue, % brown
            anchorcolor = blue]{hyperref}

\usepackage[misc]{ifsym}
\usepackage{minitoc}
\usepackage{cleveref} 
\usepackage{subcaption}
\usepackage{booktabs}
\usepackage{amsmath}
\usepackage{mathtools}
\usepackage{amssymb}
\usepackage{graphicx}   % for \resizebox
\usepackage{booktabs}   % for \toprule etc. \bottomrule
\usepackage{algorithm}
\usepackage{tcolorbox}
\usepackage{verbatim}   % for comment environment
\usepackage{makecell}   % line breaks inside cells
\usepackage{array}      % better column
\usepackage{multirow}
\usepackage{xurl}
\usepackage{amsthm}
\usepackage{booktabs}
\usepackage{adjustbox}
\usepackage{caption}

\usepackage{amssymb}% http://ctan.org/pkg/amssymb
\usepackage{pifont}% http://ctan.org/pkg/pifont
\usepackage{algorithm}% http://ctan.org/pkg/algorithm
\usepackage{algpseudocode}% http://ctan.org/pkg/algorithmicx

\usepackage{listings}
\tcbuselibrary{listings,skins,breakable}

\usepackage{tikz}
\usetikzlibrary{
    matrix,
    arrows.meta,
    positioning,
    shapes.geometric,
    fit,
    backgrounds
}

\usepackage{pdflscape}
\usepackage{afterpage}

\newcommand{\FPole}[1]{\textcolor{purple}{#1}}        % Frontal Pole
\newcommand{\Frontal}[1]{\textcolor{red}{#1}}         % Frontal
\newcommand{\Central}[1]{\textcolor{blue}{#1}}        % Central
\newcommand{\Parietal}[1]{\textcolor{orange}{#1}}     % Parietal
\newcommand{\Temporal}[1]{\textcolor{teal}{#1}}       % Temporal
\newcommand{\Occipital}[1]{\textcolor{green!60!black}{#1}} % Occipital

\makeatletter

\newcommand{\appendixtableofcontents}{%
  \begingroup
  \renewcommand{\contentsname}{Appendix Contents}%
  \@starttoc{atoc}%
  \endgroup
}

\newcommand{\startappendixtoc}{%
  \let\oldaddcontentsline\addcontentsline
  \renewcommand{\addcontentsline}[3]{%
    \def\@tempa{##1}\def\@tempb{toc}%
    \ifx\@tempa\@tempb
      \oldaddcontentsline{atoc}{##2}{##3}%
    \else
      \oldaddcontentsline{##1}{##2}{##3}%
    \fi
  }%
}
\makeatother

\title{Assessing Region-Level EEG Contributions to Cognitive Workload Prediction}

\renewcommand{\today}{}
\author[1]{Jacob Wong\textsuperscript{*}}
\author[2]{Sohan Singh\textsuperscript{*}}
\author[2]{Prannaya Gupta}
\author[3]{Jin Xing Ang}
\author[2]{Kritika Johari}
\author[2]{U-Xuan Tan}
\affil[*]{Equal Contribution}
\affil[1]{School of InfoComm Technology, Ngee Ann Polytechnic}
\affil[2]{Engineering Product Development Pillar, Singapore University of Technology and Design}
\affil[3]{NUS High School of Math and Science}

\begin{abstract}
\vspace{-0.2in}

{\fontsize{12pt}{12pt} \selectfont \raisebox{-0.06em}{\includegraphics[height=1em]{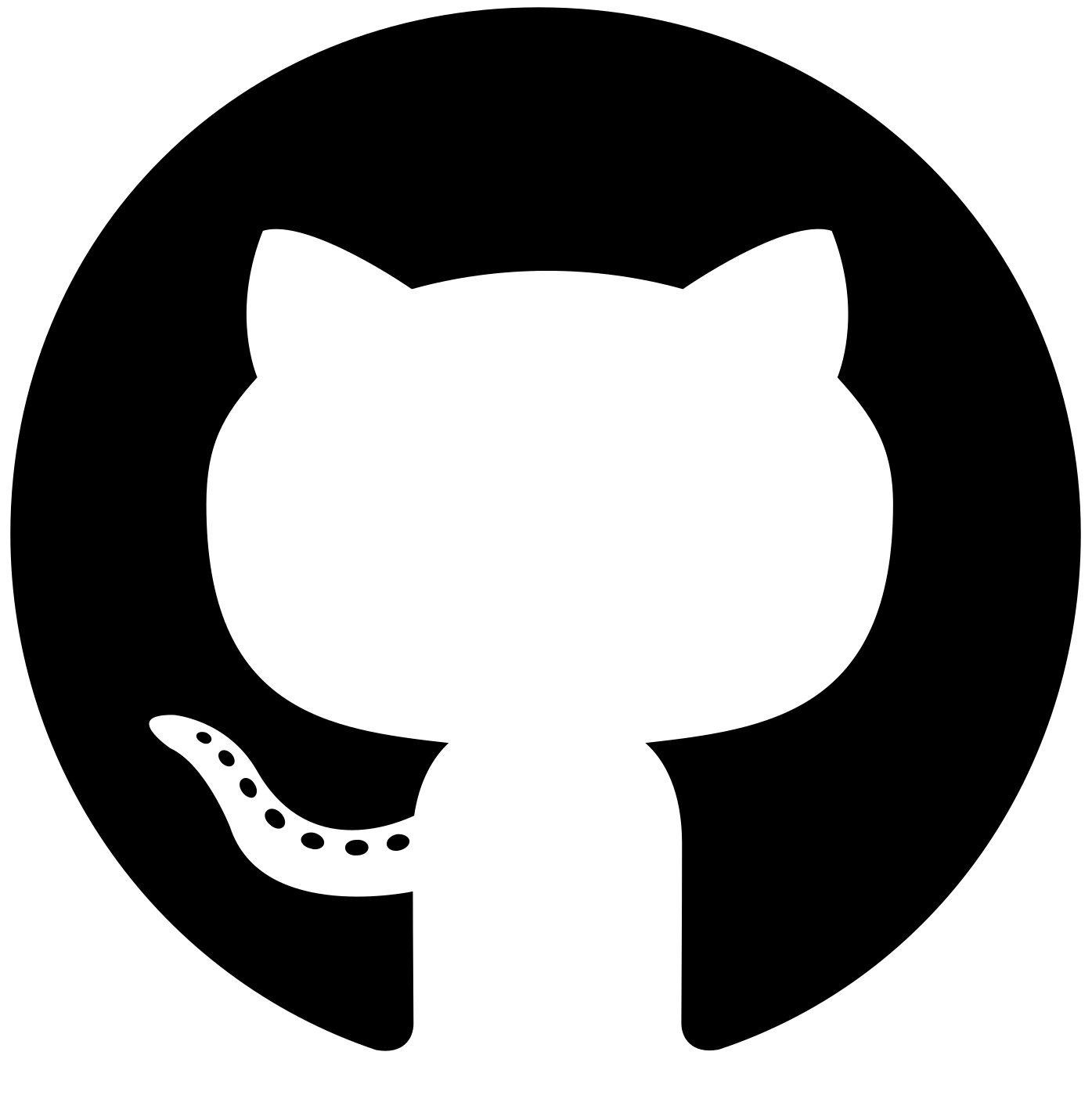}} Code: \url{https://github.com/aether-sutd/CorticalRegionSelection} %\quad \raisebox{-0.06em}{\includegraphics[height=1em]{assets/HF.png}} Models: \href{https://huggingface.co/collections/Gen-Verse/open-agentrl}{Policy \& Reward}
}
\\
\\
Accurate and generalizable estimation of cognitive workload from electroencephalography (EEG) is critical for human-centered and safety-critical systems. Although EEG is widely used for workload assessment, the consistency of region-level EEG contributions across tasks, datasets, and subjects remains unclear. This paper presents a region-level evaluation framework for EEG-based workload prediction in which models are trained and evaluated using features extracted exclusively from electrodes belonging to anatomically defined scalp regions. We perform a large-scale analysis across four publicly available EEG workload datasets spanning diverse task demands, recording hardware, and electrode montages. Region importance is quantified using a model-agnostic, performance-based approach under both mixed-subject and subject-independent evaluation protocols, with results aggregated using a rank-based strategy to ensure robustness across experimental configurations. Across all datasets and subject-independent evaluations, frontal electrode groups outperform the full-scalp baseline by approximately 15-20\% in relative rank position while using substantially fewer electrodes. Fronto-central regions exhibit the most stable predictive utility, whereas posterior and occipital regions contribute less consistently across experimental conditions. These findings indicate that workload-relevant EEG information is most consistently retained within frontal and fronto-central electrode groups, supporting the design of efficient and generalizable EEG-based workload monitoring systems.
\end{abstract}

\begin{document}

\maketitle

% \newpage
\begin{figure}[h]
    \centering
    \includegraphics[width=\linewidth]{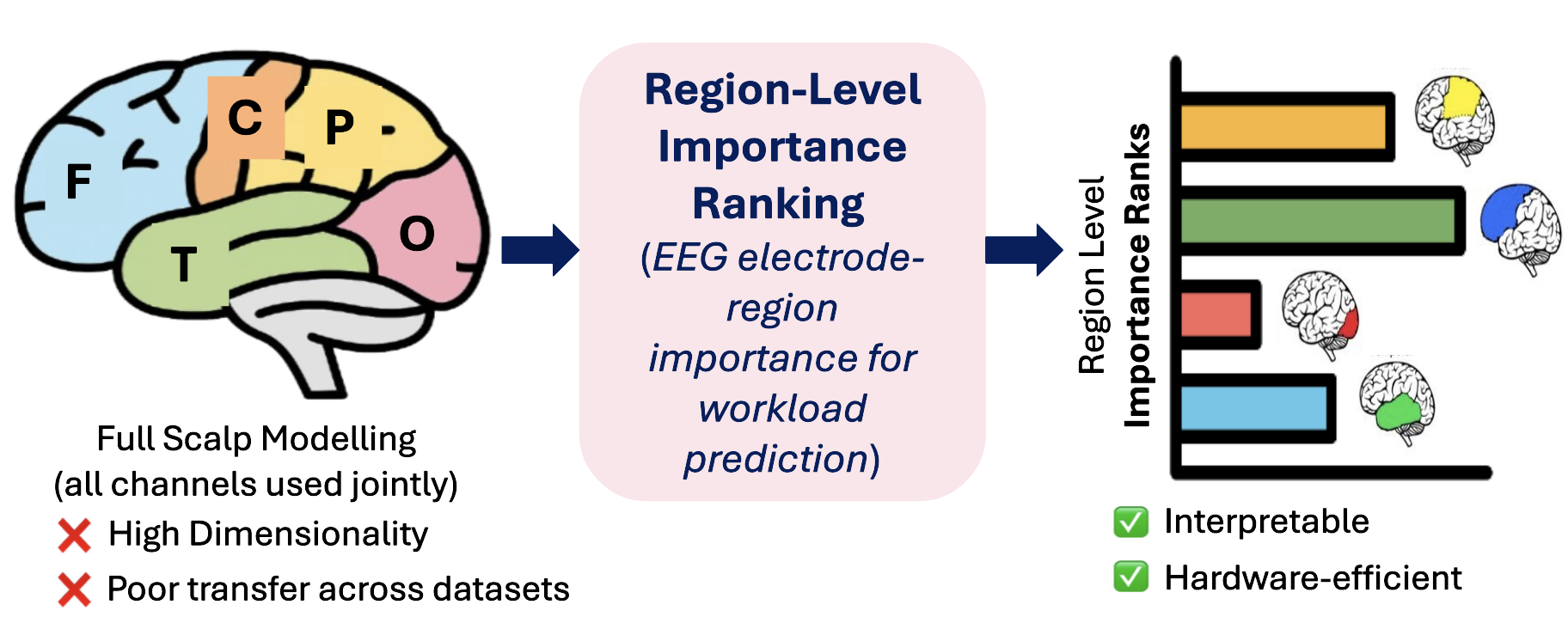}
    \caption{Conceptual overview of region-level importance analysis for EEG-based workload prediction. Instead of full-scalp modeling (left), EEG channels are grouped by anatomy and modeled independently (center), enabling region-wise importance ranking (right) with improved interpretability and hardware efficiency.}
    \label{fig:intro_fig}
\end{figure}

\section{Introduction}

Cognitive workload reflects the mental effort required to perform a task and plays a critical role in human performance, safety, and decision making~\citep{hirachan2022measuring, swerdloff2020quantifying}. Excessive workload can lead to errors, delayed responses, and reduced situational awareness~\citep{tomori2025investigating}, while insufficient workload may result in disengagement and performance decline. Reliable assessment of cognitive workload is therefore essential in human-centered and safety-critical environments~\citep{momeni2019real} such as aviation~\citep{taheri2023using}, surveillance, adaptive automation, and human--computer interaction.

Electroencephalography (EEG) provides a non-invasive and temporally precise measure of cortical activity and has been widely used to study workload-related neural dynamics~\citep{erdogmus2005cognitive, mathews2024eeg}. Prior research has shown that oscillatory activity in specific frequency bands correlates with attentional demand, working memory load, and executive control. As a result, EEG-based workload estimation has gained increasing attention for both laboratory research and real-world monitoring applications~\citep{kyriaki2024comprehensive}.

Despite recent advances, two challenges remain in EEG-based workload modeling. First, many approaches rely on full-channel montages~\citep{zhou2021cognitive}, implicitly assuming that workload-relevant information is uniformly distributed across the scalp. This increases system complexity and reduces robustness when models are transferred across datasets with different electrode configurations. Second, spatial relevance is typically evaluated within a single task or dataset. For example, \cite{mastropietro2023reliability} studied electrode configurations in one paradigm, while clinical EEG work similarly reports task-dependent channel advantages~\citep{lee2022selection}, limiting insight into cross-task generalization. These limitations are conceptually illustrated in \Cref{fig:intro_fig} (left).

More generally, prior channel-selection and feature-importance approaches are frequently model-dependent or sensitive to evaluation protocol, constraining interpretability and generalizability~\citep{prodhan2024optimal}. In this work, we address these limitations by introducing a unified, region-level evaluation framework that quantifies the predictive contribution of anatomically defined EEG electrode groups across multiple workload tasks, datasets, model families, and evaluation protocols. As shown in \Cref{fig:intro_fig}, EEG channels are grouped according to standard 10--20 and 10--10 conventions~\citep{ebersole2003current}, and region-restricted model training is used to enable consistent and leakage-free comparison of anatomically defined EEG electrode regions under heterogeneous conditions.

This work makes the following contributions:
\begin{itemize}
    \item We formulate cognitive workload prediction from EEG as a \emph{region-level inference problem}, replacing full-scalp assumptions with anatomically grounded electrode-group modeling to enable interpretable and hardware-efficient designs; this extends configuration-focused findings that electrode subsets (e.g., fronto–parietal) yield more reliable workload indices in controlled tasks~\citep{mastropietro2023reliability}.
    
    \item We develop a \emph{model-agnostic, region-restricted evaluation strategy} that quantifies the standalone predictive contribution of anatomically defined EEG electrode regions using performance degradation relative to a full-channel baseline, as conceptually summarized in \Cref{fig:intro_fig}.
    
    \item We execute a large-scale empirical study spanning multiple EEG workload datasets, feature-selection strategies, model families, and validation protocols, enabling systematic assessment of region-level robustness under heterogeneous experimental conditions.
    
    \item We derive a \emph{consensus ranking of EEG electrode regions} via rank-based aggregation and validate its stability under both mixed-subject and subject-independent evaluation, directly addressing known cross-subject generalization challenges in EEG analysis~\citep{apicella2024toward}.
\end{itemize}

\section{Methodology}

\begin{figure*}[h]
  \centering
  \includegraphics[width=\textwidth]{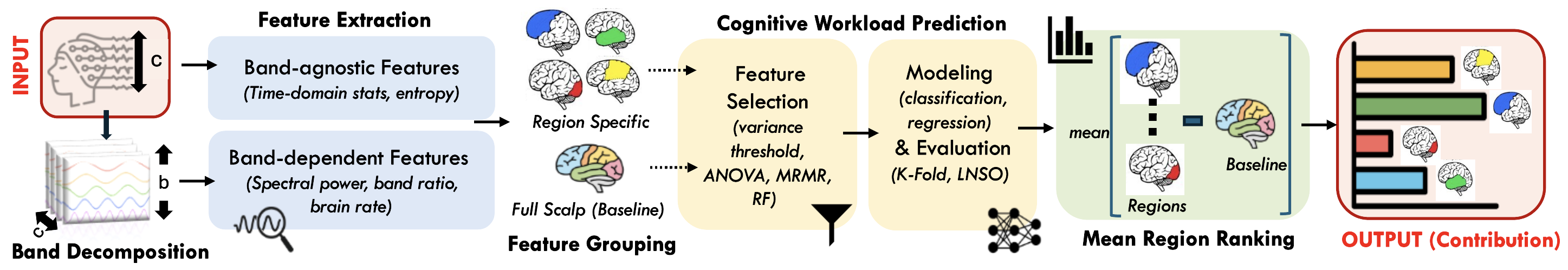}
  \caption{Overview of the proposed methodology for region-level EEG-based cognitive workload analysis. Raw EEG recordings are first processed to extract band-agnostic and band-dependent features from channel- and frequency-band representations. Extracted features are then grouped by anatomically defined EEG electrode regions, followed by region-wise feature selection and model training for workload classification or regression under both K-fold and leave-N-subjects-out (LNSO) evaluation. Region-specific model performance is compared against a full-scalp baseline, and results are aggregated across configurations to obtain a consensus ranking of EEG electrode regions based on their predictive contribution.}
  \label{fig:methodology}
\end{figure*}

We aim to quantify the relative predictive contribution of different EEG electrode regions to cognitive workload prediction in a task- and dataset-agnostic manner. Given a multichannel EEG recording collected during a workload-inducing task, our objective is to determine which anatomically defined electrode groups provide the most stable and informative neural signals across heterogeneous experimental paradigms, recording systems, and subject populations. An overview of the complete processing and analysis pipeline is shown in \Cref{fig:methodology}. The methodology comprises five main stages: EEG preprocessing and band decomposition, feature extraction, anatomical channel grouping, region-level modeling and evaluation, and aggregation of electrode-group importance across datasets.

\subsection{Datasets and Workload Label Harmonization}

\afterpage{%
    \clearpage% Flush earlier floats (otherwise order might not be correct)
    \thispagestyle{empty}% empty page style (?)
    \begin{landscape}% Landscape page

\begin{table*}[ht]
  \caption{Summary of the datasets used in this study, including participant counts, EEG acquisition devices, sampling rates, recorded channels, and workload task characteristics. Electrode colors indicate the anatomically defined EEG electrode-group assignments used for region-level analysis under the standard 10--20 and 10--10 systems: Frontal Pole (purple), Frontal (red), Central (blue), Temporal (teal), Parietal (orange), and Occipital (green).}
  \label{tab:datasets}
  \begin{tabular}{p{2cm} c p{2.5cm} c p{5cm} p{6.5cm}}
    \toprule
    \textbf{Dataset} & \textbf{No. Subjects} & \textbf{EEG Device} & \textbf{Sampling Rate (Hz)} & \textbf{Channels} & \textbf{Task Description} \\
    \midrule

    \textbf{FDE--HTC} 
    & 17 
    & Emotiv Epoc X 
    & 128 
    & \FPole{AF3}, \FPole{AF4}, \Frontal{F3}, \Frontal{F4}, \Frontal{F8}, \Frontal{FC5}, \Frontal{FC6}, 
      \Temporal{T7}, \Temporal{T8}, \Parietal{P7}, \Parietal{P8}, \Occipital{O1}, \Occipital{O2}
    & \textit{Heat-the-Chair} rapid-response task designed to induce time pressure and performance-related workload. \\
    \addlinespace

    \textbf{FDE--Nback} 
    & 17 
    & Emotiv Epoc X 
    & 128 
    & \FPole{AF3}, \FPole{AF4}, \Frontal{F3}, \Frontal{F4}, \Frontal{F8}, \Frontal{FC5}, \Frontal{FC6}, 
      \Temporal{T7}, \Temporal{T8}, \Parietal{P7}, \Parietal{P8}, \Occipital{O1}, \Occipital{O2}
    & N-back working-memory task with varying memory load to elicit sustained cognitive workload. \\
    \addlinespace

    \textbf{MOCAS} 
    & 21 
    & Emotiv Insight 
    & 128 
    & \FPole{AF3}, \FPole{AF4}, \Temporal{T7}, \Temporal{T8}, \Parietal{Pz}
    & CCTV-surveillance task involving monitoring multiple robot-controlled video feeds and identifying abnormal objects under varying workload. \\
    \addlinespace

    \textbf{WAUC} 
    & 48 
    & Neuroelectrics Enobio 
    & 500 
    & \FPole{AF7}, \FPole{Fp1}, \FPole{Fp2}, \FPole{AF8}, \Temporal{T9}, \Temporal{T10}, \Parietal{P3}, \Parietal{P4}
    & MATB-II multitasking battery including system monitoring, tracking, and resource management with manipulated workload and physical activity. \\
    \addlinespace

    \textbf{HCI-SENSE-42} 
    & 42 
    & BioSemi \newline ActiveTwo 
    & 1024 
    & \FPole{Fp1}, \FPole{Fp2}, \FPole{AF3}, \FPole{AF4}, \Frontal{F7}, \Frontal{F3}, \Frontal{Fz}, \Frontal{F4}, \Frontal{F8}, \Frontal{FC5}, \Frontal{FC1}, \Frontal{FC2}, \Frontal{FC6}, \Central{C3}, \Central{Cz}, \Central{C4}, \Temporal{T7}, \Temporal{T8}, \Parietal{CP5}, \Parietal{CP1}, \Parietal{CP2}, \Parietal{CP6}, \Parietal{P7}, \Parietal{P3}, \Parietal{Pz}, \Parietal{P4}, \Parietal{P8}, \Parietal{PO3}, \Parietal{PO4}, \Occipital{O1}, \Occipital{Oz}, \Occipital{O2}

    & Two-hour naturalistic human--computer interaction session involving continuous computer use, app switching, typing, file operations, and browsing, capturing spontaneous workload fluctuations. \\
    \bottomrule
  \end{tabular}
\end{table*}

    \end{landscape}
    \clearpage% Flush page
}

We evaluate the proposed framework using five publicly available EEG datasets encompassing diverse workload-inducing tasks, including time-pressured response (FDE--HTC;~\citealp{hernandez2024eeg}), working-memory manipulation (FDE--Nback;~\citealp{hernandez2024eeg}), visual surveillance (MOCAS;~\citealp{jo2024mocas}), multitasking and resource management (WAUC;~\citealp{albuquerque2020wauc}), and prolonged naturalistic human--computer interaction (HCI-SENSE-42;~\citealp{zhang2025multimodal}). Although these tasks differ in structure and ecological realism, all elicit cognitive workload through sustained attention, executive control, or decision-making demands. \Cref{tab:datasets} summarizes the datasets, including participant counts, EEG devices, sampling rates, channel configurations, and task descriptions. The diversity of task paradigms and acquisition hardware motivates a region-based analysis that enables consistent comparison across heterogeneous experimental settings. To ensure signal-level consistency across datasets recorded at different sampling rates, all EEG recordings were resampled to a common rate of 128~Hz using polyphase filtering with anti-aliasing. Resampling was applied independently to each channel, after which frequency-band decomposition was recomputed to maintain consistent band definitions.

Cognitive workload labels in all datasets were derived from the NASA Task Load Index (NASA-TLX;~\citealp{hart1988development}), a widely used and validated subjective workload assessment tool. NASA-TLX is specifically designed to capture perceived task demand across multiple dimensions, including mental effort, time pressure, and frustration, and has been extensively used as supervision in EEG-based workload modeling~\citep{zenia2025eeg, hirachan2022measuring}. Using NASA-TLX therefore aligns our analysis with established practice in the literature while directly targeting human-experienced workload~\citep{mohanavelu2020dynamic}. Because NASA-TLX scores were originally reported on different scales across datasets, all workload values were linearly rescaled to a common 0--100 range using min--max normalization,
\[
\text{Workload}_n = 100 \times \frac{\text{score} - \text{score}_{\min}}{\text{score}_{\max} - \text{score}_{\min}} .
\]

This transformation preserves the relative spacing of subjective ratings while enabling cross-dataset comparability. For classification experiments, the normalized workload scores were further discretized into three workload categories (low, medium, and high) using dataset-specific empirical 33rd and 67th percentile thresholds computed independently for each dataset. This percentile-based discretization adapts to differences in workload-score distributions across datasets while reducing sensitivity to small subjective score variations.

\begin{algorithm}[t]
\caption{EEG Signal Preprocessing Pipeline following standard EEG band decomposition and feature-extraction.}

\label{alg:eeg_preproc}
\begin{algorithmic}[1]

\State \textbf{Input:} Raw EEG recording 
\[
X(t) = [x_1(t), x_2(t), \dots, x_C(t)]
\]

\State \textbf{Output:} Sub-band signals 
\[
\{x_{c,b}(t)\;|\; c = 1,\dots,C;\; b \in \{\delta,\theta,\alpha,\beta,\gamma\}\}
\]

\Statex
\State \textbf{Step 1: Broadband Pre-filtering (0.5--45 Hz)}
\For{$c = 1$ to $C$}
    \State Apply a 0.5--45 Hz Butterworth bandpass filter to $x_c(t)$
    \State Store the filtered signal as $\tilde{x}_c(t)$
\EndFor

\Statex
\State \textbf{Step 2: Frequency-Band Decomposition}
\For{$c = 1$ to $C$}
    \For{$b \in \{\delta,\theta,\alpha,\beta,\gamma\}$}
        \State Apply the bandpass filter for band $b$ to $\tilde{x}_c(t)$
        \State Obtain sub-band signal $x_{c,b}(t)$
    \EndFor
\EndFor

\Statex
\State \Return $\{x_{c,b}(t)\}_{c,b}$

\end{algorithmic}
\end{algorithm}

\subsection{EEG Preprocessing and Frequency-Band Decomposition}

The EEG preprocessing and band-decomposition procedure used in this paper is summarized in \Cref{alg:eeg_preproc}. To remove non-neural artifacts and isolate physiologically meaningful oscillatory activity, each EEG channel was first bandpass filtered between 0.5 and 45~Hz using a fourth-order Butterworth IIR filter~\citep{cohen2014analyzing}. This broadband pre-filtering attenuates slow baseline drift, power-line interference, and high-frequency muscle artifacts, yielding a cleaned signal suitable for spectral analysis.

Following broadband denoising, each channel was decomposed into the standard EEG frequency bands: delta (0.5--4~Hz), theta (4--8~Hz), alpha (8--13~Hz), beta (13--30~Hz), and gamma (30--45~Hz)~\citep{niedermeyer2005electroencephalography}. As detailed in \Cref{alg:eeg_preproc}, band-specific signals were obtained by applying bandpass filters to the denoised channel signal. This decomposition isolates distinct oscillatory components associated with cognitive workload and enables subsequent band-specific feature extraction~\citep{klimesch1999eeg}.

\subsection{Feature Extraction}

Features were extracted independently from each band-limited signal for every channel. We compute a set of interpretable spectral, statistical, and entropy-based features commonly used in EEG workload analysis~\citep{stancin2021review}.

\paragraph{Band-Specific Power Features.}
For each channel $c$ and frequency band $b$, we compute the absolute band power $P_b^{(c)}$, defined as the total spectral power within the frequency range of band $b$ \citep{cohen2014analyzing}. To reduce sensitivity to inter-subject and hardware-related amplitude differences, relative band power is computed as
\[
p_b^{(c)} = \frac{P_b^{(c)}}{P_{\text{signal}}^{(c)}},
\]
where $P_{\text{signal}}^{(c)}$ denotes the total broadband power of channel $c$ over 0.5--45~Hz.

\paragraph{Band Power Ratios and Brain Rate.}
To capture interactions between oscillatory components, we compute a subset of commonly used band-power ratios (e.g., $\theta/\beta$, $(\theta+\alpha)/(\alpha+\beta)$). In addition, we compute the brain rate (BR;~\citealp{longo2022modeling}, a spectrum-weighted summary measure defined as
\[
\text{BR}(c) = \sum_{b \in \{\delta,\theta,\alpha,\beta,\gamma\}} f_b \cdot p_b^{(c)},
\]
where $f_b$ denotes the mean frequency of band $b$. BR provides a compact descriptor of the distribution of oscillatory activity across frequency bands.

\paragraph{Statistical and Entropy Features.}
For each band-limited signal, we extract time-domain statistical features (mean, variance, skewness, kurtosis, zero-crossing rate) and entropy-based measures (Rényi entropy, differential entropy, and fuzzy entropy), capturing signal variability and complexity~\citep{stancin2021review}.

\subsection{Region-Level Modeling and Importance Analysis}

To enable region-level analysis of EEG electrode-region contributions to cognitive workload prediction, EEG-derived features were grouped according to the anatomical locations of their corresponding electrodes, as illustrated in \Cref{fig:methodology}. Channel assignments follow the international 10--20 and 10--10 placement systems, in which electrode labels encode approximate cortical location~\citep{ebersole2003current}. The channel-to-region mappings used in this study are explicitly indicated in \Cref{tab:datasets}. Six primary anatomical regions were considered: Frontal Pole, Frontal, Temporal, Central, Parietal, and Occipital~\citep{mastropietro2023reliability}. To capture broader spatial patterns and ensure robustness across datasets with heterogeneous electrode montages, extended composite regions were also defined by combining adjacent primary regions (e.g., Frontal-Extended, Posterior, Central-Extended, Temporal-Extended). These extended regions were designed to capture spatially distributed EEG representations that may span neighboring anatomical areas and therefore not be fully isolated within narrowly localized electrode groups. This formulation also improves robustness to sparse or heterogeneous electrode configurations while preserving anatomical interpretability. Primary regions represent localized electrode groupings, whereas extended regions capture broader spatial EEG patterns across adjacent scalp areas.

To quantify region-level predictive contributions, feature selection, model training, and evaluation were conducted independently for each anatomical electrode group, following the region-restricted pipeline shown in \Cref{fig:methodology}. This processing ensures that each model is trained exclusively on features derived from a single region and prevents information leakage across regions. For each dataset, electrode group, and training fold, features with near-zero variance (standard deviation $<10^{-10}$) were first removed~\citep{kuhn2013applied}. Statistical feature selection was then applied within the training fold only, retaining the top $k=24$ features per group. Three complementary selection strategies were employed: ANOVA F-tests~\citep{guyon2003introduction}, minimum-redundancy–maximum-relevance (MRMR)~\citep{ding2005minimum}, and Random Forest feature importance~\citep{scikit-learn}.

Region-specific models were evaluated using two complementary protocols, as summarized in \Cref{fig:methodology}. Standard three-fold cross-validation estimates performance under mixed-subject conditions, while leave-$N$-subjects-out (LNSO) evaluation assesses subject-independent generalization. In LNSO, $N=\max(1,\lfloor n_{\text{subjects}}/5 \rfloor)$ subjects were held out per fold, ensuring that no participant appears in both training and test sets. To ensure that the observed region-level trends were not tied to a specific predictive architecture, experiments were conducted across multiple model families spanning both linear and nonlinear learning paradigms. Classification experiments included Logistic Regression, Support Vector Machines (SVM), Random Forest, and Gradient Boosting classifiers, while regression analyses employed Linear Regression, Support Vector Regression (SVR), Random Forest Regression, and Gradient Boosting Regression. Finally, to obtain a unified, dataset- and model-agnostic estimate of electrode-group relevance, region-level performance results were aggregated across datasets, feature-selection methods, model families, and evaluation protocols. For each configuration, electrode groups were ranked by predictive performance, and a consensus importance score was computed as the mean rank across all valid configurations.

\section{Results}
This section presents the empirical findings of our region-level EEG analysis for cognitive workload prediction. We first report global electrode-group rankings aggregated across datasets, models, and evaluation protocols. We then analyze region-level patterns across individual NASA–TLX subscales to assess consistency across workload dimensions. Finally, we examine dataset-specific deviations and importance–stability trade-offs to contextualize task-dependent effects and cross-dataset robustness.

\subsection{Global Electrode-Group Rankings}

\begin{figure}
    \centering
    \includegraphics[width=0.6\linewidth]{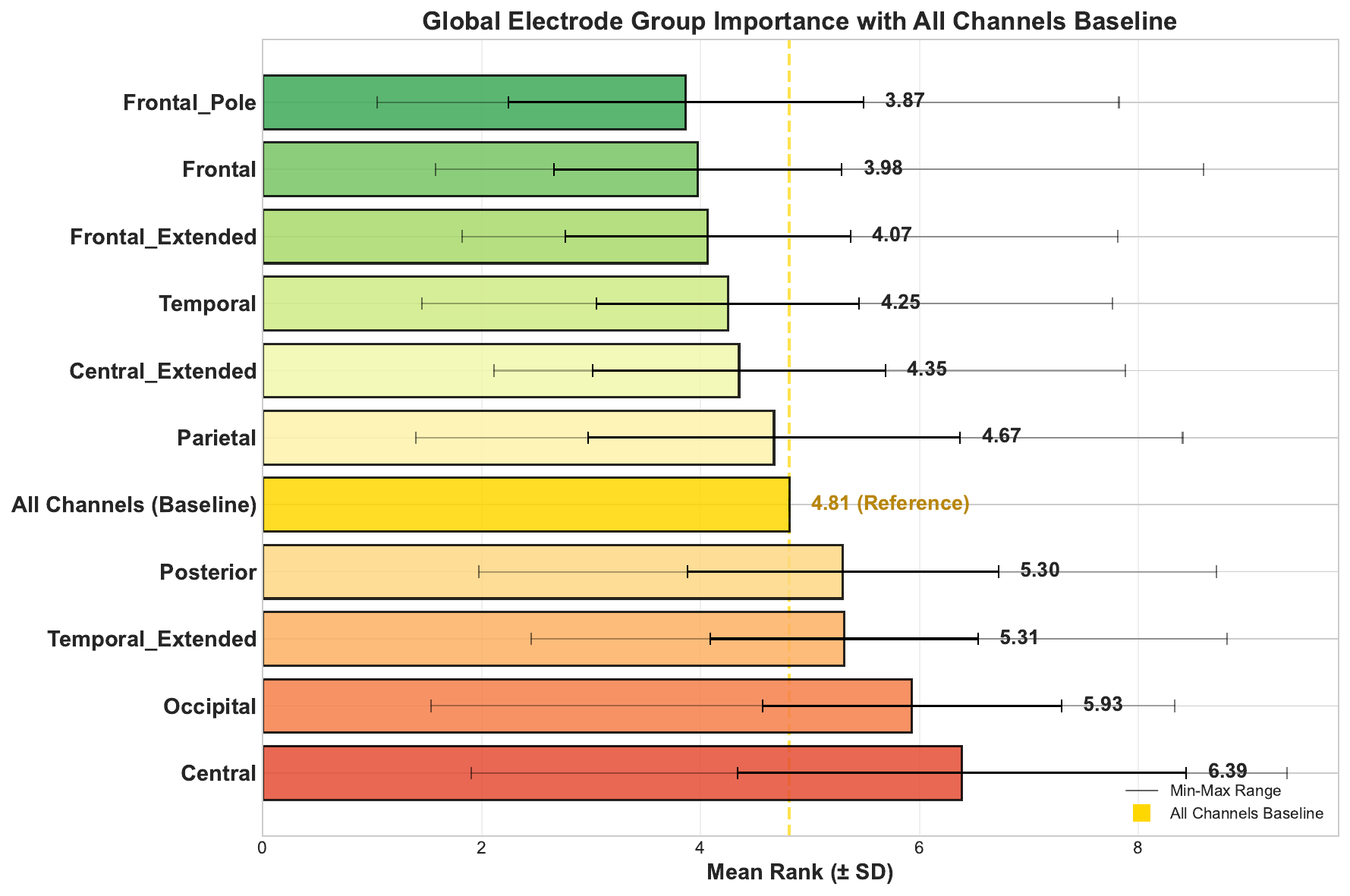}
    \caption{Global importance ranking of EEG electrode groups aggregated across all datasets, feature-selection methods, workload sub-scales, prediction tasks, and evaluation protocols. Bars indicate mean rank (lower is better), with error bars showing ± one standard deviation and whiskers denoting the minimum–maximum rank range observed across configurations. The full-scalp (all-channels) model is included as a reference baseline.}
    \label{fig:global_rankings}
\end{figure}

\begin{figure*}[!ht]
    \centering
    \includegraphics[width=\textwidth]{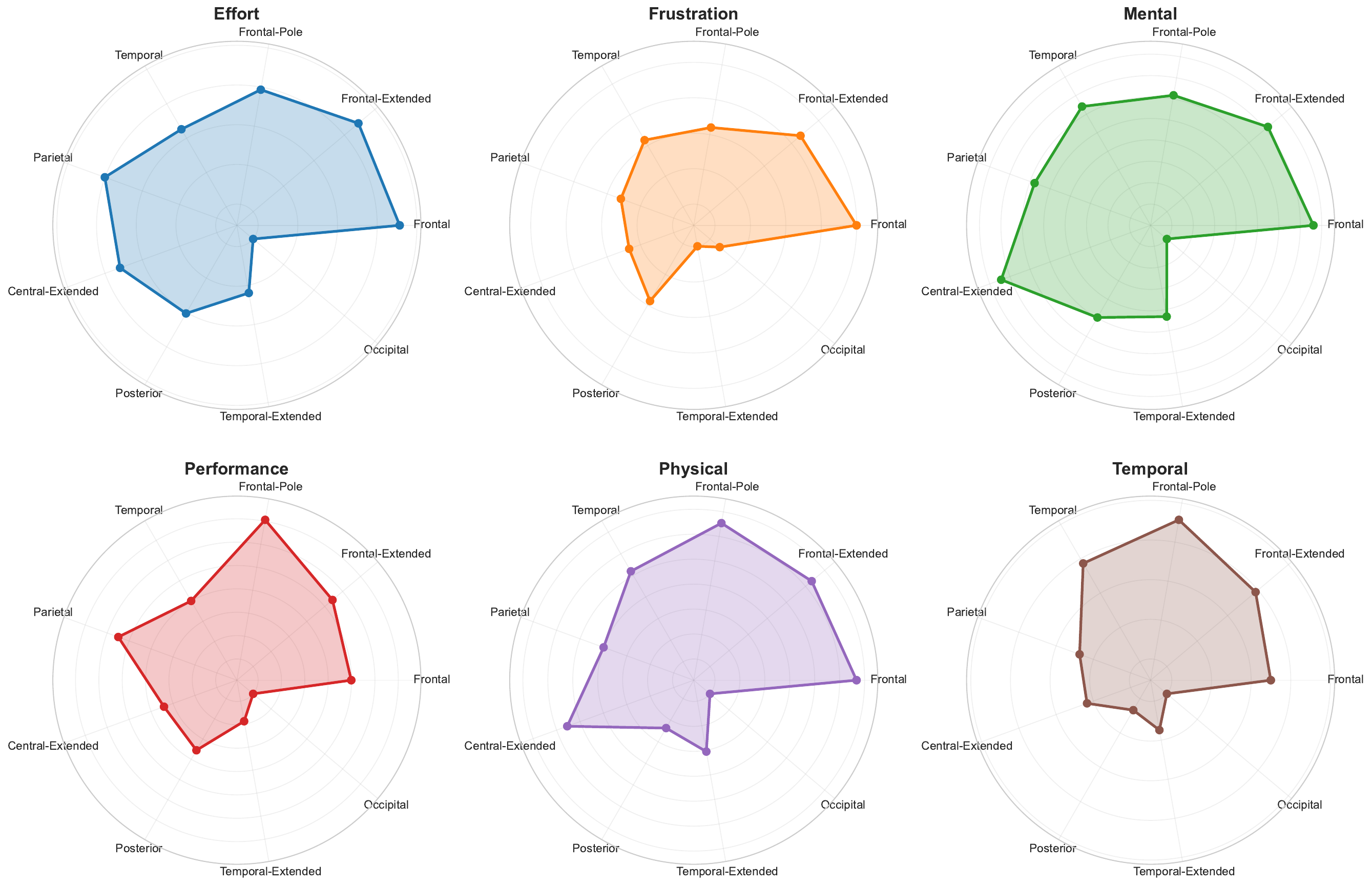}
    \caption{Radar plots showing electrode-group importance across NASA–TLX subscales based on mean rank values. Radial distance represents inverted ranking, such that electrode groups plotted farther from the center correspond to higher predictive importance (lower mean rank). Consistent outward dominance of frontal and frontal-extended regions across subscales indicates stable contribution to perceived workload dimensions.}
    \label{fig:Subscale_agreement}
\end{figure*}

\Cref{fig:global_rankings} summarizes the global importance of each electrode group, aggregated across all datasets, feature-selection methods, workload sub-scales, prediction tasks, and evaluation protocols. Lower mean ranks indicate higher predictive importance. Error bars denote the standard deviation across configurations, while whiskers show the minimum and maximum ranks observed. An all-channels baseline is included for reference.

A clear spatial hierarchy emerges. Frontal-dominant regions consistently achieve the lowest mean ranks, indicating the strongest and most reliable predictive contribution. In particular, the Frontal, Frontal Pole, and Frontal-Extended groups outperform the all-channels baseline, despite using a reduced subset of electrodes. This demonstrates that workload-relevant information is not uniformly distributed across the scalp and that restricting analysis to anatomically informative regions can improve predictive focus rather than degrade performance.

Temporal and Central-Extended regions form a secondary tier, achieving moderate mean ranks with relatively low variability. These regions appear to capture task-relevant signals in a stable but less dominant manner, likely reflecting supporting sensorimotor and temporal processing demands.

In contrast, Posterior, Occipital, and strict Central groups consistently rank below the baseline. Their higher mean ranks indicate weaker predictive contribution when evaluated in isolation, suggesting that these regions encode either peripheral or task-specific information that does not generalize robustly across datasets. Overall, \Cref{fig:global_rankings} shows that workload-related EEG information is
concentrated in frontal and fronto-central regions, while posterior and occipital
areas contribute less consistently under heterogeneous conditions, consistent
with prior electrode-configuration reliability findings~\citep{mastropietro2023reliability}.

\subsection{Subscale-Specific Region-Level EEG Contributions}

While overall workload scores provide a global measure of task demand, the NASA–TLX framework decomposes workload into multiple subscales reflecting distinct cognitive, affective, and temporal components. Examining electrode-group importance at the subscale level allows us to assess whether region-level EEG contributions are driven by specific workload dimensions or remain stable across different aspects of perceived task demand.

\Cref{fig:Subscale_agreement} shows radar plots of electrode-group importance across the six NASA–TLX subscales using mean rank values. Radial distance encodes inverted rank, such that electrode groups plotted farther from the center indicate higher predictive importance. The strong visual overlap across subscales indicates that the spatial distribution of workload-related EEG information remains largely consistent regardless of the specific workload dimension considered.

Across the Effort, Frustration, Mental, and Physical subscales, the Frontal and Frontal-Extended regions consistently appear farthest from the center, indicating high and stable importance. This pattern highlights the central role of frontal activity in capturing cognitive demand, affective strain, and perceived exertion, consistent with established links between frontal networks and executive control, attentional regulation, and resource allocation during demanding tasks.

A modest shift in regional emphasis is observed for the Performance and Temporal subscales, where Frontal Pole regions exhibit relatively greater radial extent. These dimensions emphasize self-evaluation, time pressure, and pacing, which rely more strongly on reflective judgment and internal monitoring processes commonly associated with anterior prefrontal cortex.

Taken together, the radar plots reveal gradual variations in regional emphasis rather than abrupt reordering of electrode-group importance across subscales. This consistency suggests that NASA–TLX dimensions reflect overlapping neural processes supported by distributed cortical networks, with frontal regions providing a stable and informative core across all components of perceived workload.

\subsection{Dataset-Specific Regional Importance Patterns}

\begin{figure}[!ht]
    \centering
    \includegraphics[width=0.6\linewidth]{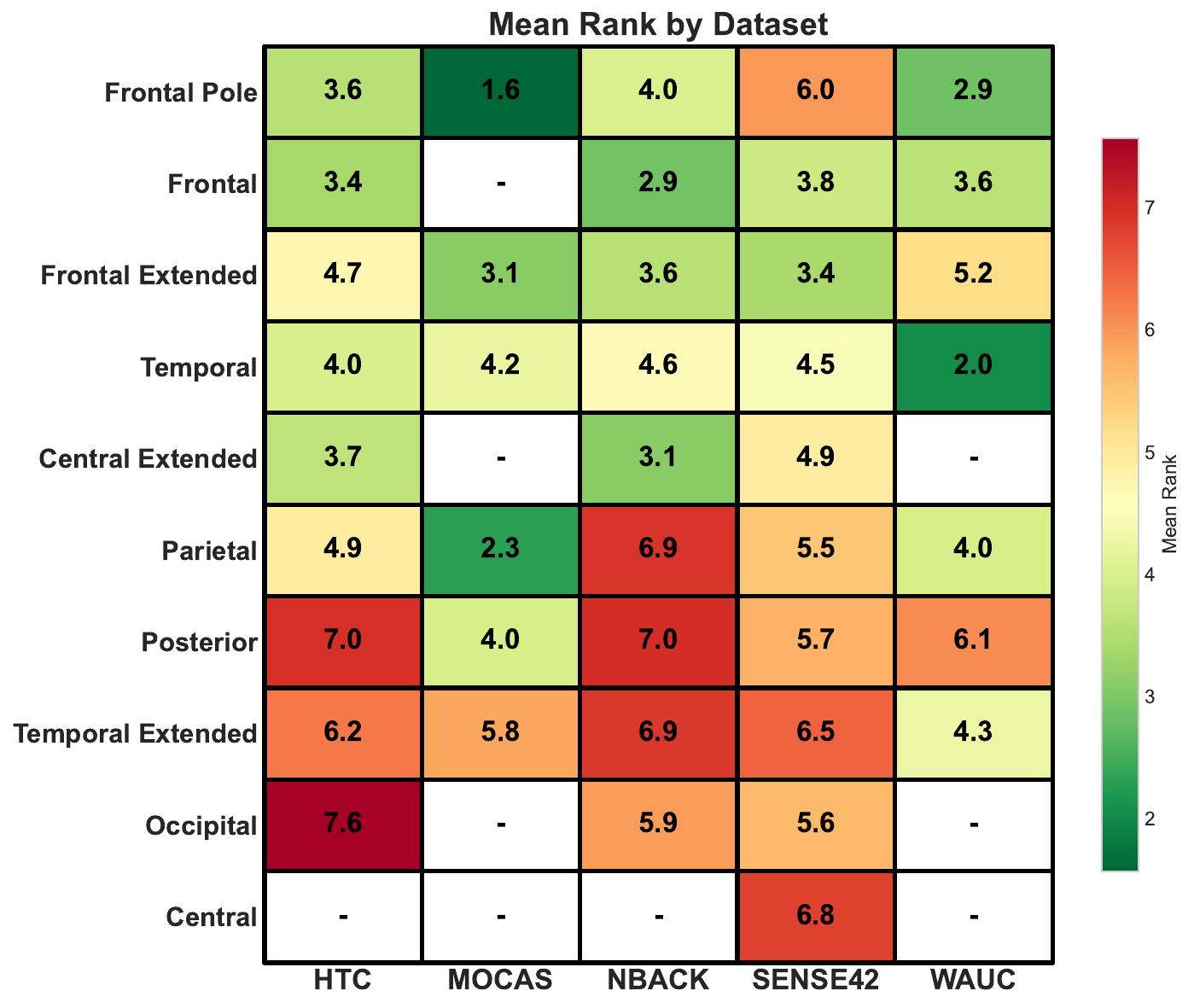}
    \caption{Dataset-specific agreement of electrode-group importance. Heatmap of mean ranking values for each electrode group across datasets/tasks: HTC (time-pressured response), MOCAS (visual surveillance), N-back (working memory), SENSE-42 (naturalistic HCI), and WAUC (multitasking). Lower ranks indicate higher predictive importance. Missing entries correspond to electrode groups not present in a given dataset. }
    \label{fig:electrode_groups_heatmap}
\end{figure}

While global rankings reveal stable trends, dataset-level analysis highlights task-dependent modulation of regional importance. \Cref{fig:electrode_groups_heatmap} shows mean electrode-group rankings for each dataset.

Across all tasks, frontal and frontal-extended regions consistently ranked among the most informative. However, specific deviations were observed. In the HCI-SENSE-42 dataset, Frontal Pole electrodes exhibited reduced importance, consistent with increased susceptibility to eye movements and facial activity during prolonged naturalistic interaction. In contrast, the WAUC dataset showed relatively stronger temporal involvement, likely reflecting task-correlated physiological and motion-related activity associated with physical workload components.

Despite these variations, no dataset overturned the broader frontal–central dominance, reinforcing the cross-task robustness of these regions for workload estimation.

\section{Conclusion}

This paper presented a region-level framework for evaluating the predictive contribution of EEG electrode regions to cognitive workload prediction across heterogeneous datasets, tasks, and evaluation protocols. By training models on anatomically defined electrode groups, we showed that workload-relevant EEG information is most consistently retained within frontal and fronto-central regions. Across five datasets and subject-independent evaluations, these regions outperformed full-scalp baselines while requiring substantially fewer electrodes, demonstrating improved efficiency and robustness for practical deployment.

Subscale-specific analysis further revealed stable frontal dominance across multiple dimensions of perceived workload. Frontal and frontal-extended regions ranked highest for effort, mental demand, frustration, and physical demand, while frontal-pole regions contributed more strongly to performance and temporal demand. The consistency of region-level rankings across diverse workload paradigms suggests that the identified EEG representations capture generalizable markers of perceived workload relevant to real-world monitoring and adaptive system design. Future work will extend this framework to multimodal sensing and real-time adaptive interfaces.

\section{Limitations and Future Work}

This study models cognitive workload using NASA--TLX scores, which reflect perceived rather than purely objective neural load. While NASA--TLX is specifically designed to capture human-experienced task demand, subjective self-report inevitably introduces inter-individual variability and label noise, potentially limiting the upper bound of achievable prediction performance. As a result, the reported findings should be interpreted as estimates of perceived workload rather than absolute cognitive load. Future work will explore the integration of objective or multimodal workload indicators to complement subjective labels and improve robustness. The datasets analyzed in this work emphasize executive control, working memory, sustained attention, multitasking, and prolonged human--computer interaction. Consequently, the observed frontal and fronto-central dominance reflects workload patterns within these operationally relevant task classes. Consistent with prior findings, cortical region relevance depends on both task demands and electrode configuration; for example, \cite{mastropietro2023reliability} reports that fronto--parietal montages yield the most reliable workload estimates in controlled tasks, whereas minimal configurations exhibit reduced stability. Accordingly, the region rankings reported here should be interpreted as task-general within operational workload contexts rather than universally optimal across all cognitive paradigms. Future studies will extend this framework to additional task domains and electrode configurations to further assess the generality of region-level workload encoding.

\section*{Acknowledgment}

The authors thank the creators and maintainers of the publicly available EEG datasets used in this study for making their data openly accessible. We also acknowledge the open-source software community for providing the tools that enabled reproducible data processing and analysis.

\bibliography{main}

% \clearpage

% \appendix

% \section*{Appendix}

% \startappendixtoc        % 这句一定要在 appendix 的 section 出现前
% \setcounter{tocdepth}{2} % 控制 appendix 目录深度（可选）

% \appendixtableofcontents % 打印 Appendix Contents（只会包含 appendix）

\end{document}